\documentclass[10pt,twocolumn,letterpaper]{article}

\usepackage{iccv}
\usepackage{times}
\usepackage{epsfig}
\usepackage{graphicx}
\usepackage{amsmath}
\usepackage{amssymb}
\usepackage{multirow}
\usepackage{subcaption}
\usepackage{algorithm}
\usepackage{algpseudocode}
\usepackage{enumitem}

\usepackage[breaklinks=true,bookmarks=false]{hyperref}

\iccvfinalcopy 

\def\iccvPaperID{****} 

\ificcvfinal\fi

\begin{document}

\title{Admix: Enhancing the Transferability of Adversarial Attacks}


\author{Xiaosen Wang\textsuperscript{\rm 1} \quad Xuanran He\textsuperscript{\rm 2} \quad Jingdong Wang\textsuperscript{\rm 3} \quad Kun He\textsuperscript{\rm 1}\thanks{Corresponding author.}\\
\textsuperscript{\rm 1}School of Computer Science and Technology, Huazhong University of Science and Technology\\
\textsuperscript{\rm 2}Wee Kim Wee School of Communication and Information, Nanyang Technological University\\
\textsuperscript{\rm 3}Microsoft Research Asia\\
{\tt\small xiaosen@hust.edu.cn, xhe015@e.ntu.edu.sg, jingdw@microsoft.com, brooklet60@hust.edu.cn}
}

\maketitle
\ificcvfinal\thispagestyle{empty}\fi

\begin{abstract}

Deep neural networks are known to be extremely vulnerable to adversarial examples under white-box setting. Moreover, the malicious adversaries crafted on the surrogate (source) model often exhibit black-box transferability on other models with the same learning task but having different architectures. Recently, various methods are proposed to boost the adversarial transferability, among which the input transformation is one of the most effective approaches. We investigate in this direction and observe that existing transformations are all applied on a single image, which might limit the adversarial transferability. To this end, we propose a new input transformation based attack method called \textit{Admix} that considers the input image and a set of images randomly sampled from other categories. Instead of directly calculating the gradient on the original input, \textit{Admix} calculates the gradient on the input image admixed with a small portion of each add-in image while using the original label of the input to craft more transferable adversaries.
Empirical evaluations on standard ImageNet dataset demonstrate that \textit{Admix} could achieve significantly better transferability than existing input transformation methods under both single model setting and ensemble-model setting. By incorporating with existing input transformations, our method could further improve the transferability and outperforms the state-of-the-art combination of input transformations by a clear margin when attacking nine advanced defense models under ensemble-model setting. Code is available at \url{https://github.com/JHL-HUST/Admix}.

\end{abstract}

\section{Introduction}

A great number of works~\cite{goodfellow2015FGSM,carlini2017cw,athalye2018obfuscated} have shown that deep neural networks (DNNs) are vulnerable to adversarial examples~\cite{szegedy2014intriguing,goodfellow2015FGSM}, \ie the malicious crafted inputs that are indistinguishable from the legitimate ones but can induce misclassification on the deep learning models. 
Such vulnerability poses potential threats to security-sensitive applications, \eg face verification~\cite{sharif2016accessorize}, autonomous driving~\cite{Eykholt2018Robust} and has inspired a sizable body of research on adversarial attacks~\cite{moosavi2016deepfool,carlini2017cw,madry2018pgd,dong2018boosting,li2019nattack,dong2019evading,xie2019improving,lin2020nesterov}. 
Moreover, the adversaries often exhibit transferability across neural network models~\cite{papernot2017practical}, in which the adversarial examples generated on one model may also mislead other models.
The adversarial transferability matters because hackers may attack a real-world DNN application without knowing any information of the target model. However, under white-box setting where the attacker has complete knowledge of the target model, existing attacks~\cite{carlini2017cw,kurakin2017IFGSM,athalye2018obfuscated,madry2018pgd} have demonstrated great attack performance but with comparatively low transferability against models with defense mechanisms~\cite{madry2018pgd, tramer2018ensemble}, making it inefficient for real-world adversarial attacks.  

To improve the transferability of adversarial attacks, various techniques have been proposed, such as advanced gradient calculations~\cite{dong2018boosting,lin2020nesterov,wang2021Enhancing}, ensemble-model attacks~\cite{liu2017delving,Li20AAAI}, input transformations~\cite{xie2019improving,dong2019evading,lin2020nesterov,zou2020improving} and model-specific methods~\cite{Wu2020Skip}. 
The input transformation (\eg randomly resizing and padding, translation, scale \etc) is one of the most effective approaches. Nevertheless, we observe that existing methods are all applied on a single input image. 
Since adversarial attacks aim to mislead the DNNs to classify the adversary into other categories, it naturally inspires us to explore whether we could further enhance the transferability by incorporating the information from other categories.

The \textit{mixup} operation, that linearly interpolates two random images and corresponding labels, is firstly proposed as a data augmentation approach to improve the generalization of standard training \cite{zhang2018mixup, verma2019manifold, yun2019cutmix}. 
Recently, \textit{mixup} is also used for inference~\cite{pang2020mixup} or adversarial training~\cite{lamb2019interpolated,lee2020adversarial} to enhance the model robustness. 
Since \textit{mixup} adopts the information of a randomly picked image, we try to directly adopt \textit{mixup} to craft adversaries but find that 
the attack performance decays significantly under white-box setting with little improvement on transferability. To craft highly transferable adversaries with the information from other categories but not harm the white-box attack performance, we propose a novel attack method called \textit{Admix} that calculates the gradient on the admixed image combined with the original input and images randomly picked from other categories.
Unlike \textit{mixup} that treats the two images equally and mixes their labels accordingly, the \textit{admix} operation adds a small portion of the add-in image from other categories to the original input but does not change the label. Thus \textit{Admix} attack could obtain diverse inputs for gradient calculation.

Empirical evaluations on standard ImageNet dataset \cite{russakovsky2015imagenet} demonstrate that, compared with existing input transformations~\cite{xie2019improving,dong2019evading,lin2020nesterov}, the proposed \textit{Admix} attack achieves significantly higher attack success rates under black-box setting and maintains similar attack performance under white-box setting. 
By incorporating \textit{Admix} with other input transformations, the transferability of the crafted adversaries could be further improved. Besides, the evaluation of the integrated method under the ensemble-model setting~\cite{liu2017delving} against nine advanced defense methods ~\cite{liao2018defense,xie2018mitigating,xu2018BitReduction,liu2019FD,guo2018countering,cohen2019certified,salman2019provably,naseer2020NRP} demonstrates that the final integrated method, termed \textit{Admix}-TI-DIM, outperforms the state-of-the-art SI-TI-DIM~\cite{lin2020nesterov} by a clear margin of 3.4\% on average, which further demonstrates the high effectiveness of \textit{Admix}.

\section{Related Work}
In this section, we provide a brief overview of the adversarial attack methods and the \textit{mixup} family.

\subsection{Adversarial Attacks and Transferability}
According to the threat model, existing attack methods can be categorized into two settings: a) white-box attack has full knowledge of the threat model, \eg(hyper-)parameters, gradient, architecture, \etc. b) black-box attack could only access to the model outputs or nothing about the threat model. In this work, we mainly focus on generating highly transferable adversaries without any knowledge of the target model, falling into the black-box setting.

Szegedy \etal~\cite{szegedy2014intriguing} first point out the existence of adversarial examples for DNNs and propose a box-constrained L-BFGS method to find adversarial examples. 
To accelerate the adversary generation process, Goodfellow \etal~\cite{goodfellow2015FGSM} propose fast gradient sign method (FGSM) to generate adversarial examples with one step of  gradient update. Kurakin \etal~\cite{kurakin2017IFGSM} further extend FGSM to an iterative version denoted as I-FGSM that exhibits higher attack success rates. Carlini \etal~\cite{carlini2017cw} propose a powerful optimization-based method by optimizing the distance between an adversary and the corresponding benign example. Though the above attacks have achieved remarkable attack performance under white-box setting, they often exhibit weak transferability.

Recently, some works focus on generating more transferable adversarial examples, which could be roughly split into four categories, namely ensemble-model attack, momentum-based attack, input transformation based attack and model-specific attack. Liu  \etal~\cite{liu2017delving} first propose an ensemble-model attack that attacks multiple models simultaneously so as to enhance the transferability. Li \etal \cite{Li20AAAI} generate adversarial examples on multiple ghost networks obtained by perturbing the dropout and skip-connection layer. Some works focus on advanced gradient calculation that adopts momentum to generate more transferable adversaries. Dong \etal~\cite{dong2018boosting} integrate momentum into I-FGSM denoted as MI-FGSM and Lin \etal~\cite{lin2020nesterov} adopt Nesterov's accelerated gradient, to further enhance the transferability. 

Several input transformation methods have also been proposed to promote the transferability.
Xie \etal~\cite{xie2019improving} propose to adopt diverse input pattern by randomly resizing and padding for gradient calculation. Dong \etal~\cite{dong2019evading} convolve the gradient with a pre-defined kernel which leads to higher transferability against models with defense mechanism. Lin \etal~\cite{lin2020nesterov} calculate the gradient on a set of scaled images to enhance the transferability. Zou \etal~\cite{zou2020improving} propose a three-stage pipeline to generate more transferable adversarial examples, namely resized-diverse-inputs, diversity-ensemble and region fitting. Wu \etal~\cite{Wu2020Skip} find that utilizing more gradient of skip connections rather than the residual modules in ResNet~\cite{he2016resnet} could enhance the transferability.




Note that the ensemble-model attack, momentum based attack, and input transformation based attack could be integrated with each other to achieve higher transferability. The proposed \textit{Admix} falls into the input transformation category, and \textit{Admix} can be combined with other input transformations as well as the other two types of attacks to further boost the transferability. 

\begin{table*}
\begin{center}
\scalebox{0.9}{
\begin{tabular}{|c|ccccccc|}
\hline
Attack & Inc-v3 & Inc-v4 & IncRes-v2 & Res-101 & Inc-v3$_{ens3}$ & Inc-v3$_{ens4}$ & IncRes-v2$_{ens}$\\
\hline\hline
MI-FGSM & 100.0 & 43.6 & 42.4 & 35.7 & 13.1 & 12.8 &  6.2\\
\textit{Mixup} & ~~71.8 & 44.2 & 41.1 & 39.0 & 13.5 & 13.4 & 7.2\\
\hline
\end{tabular}
}
\vspace{-0.3em}
\caption{Attack success rates (\%) of MI-FGSM and \textit{mixup} transformation. The adversaries are crafted on Inc-v3 model.}
\label{tab:mixup}
\end{center} 
\vspace{-1em}
\end{table*}
\subsection{The Mixup Family}
Zhang \etal \cite{zhang2018mixup} first propose a novel method called \textit{mixup} to improve the model generalization by interpolating two randomly sampled examples $(x, y)$ and $(x', y')$ with $\lambda \in [0,1]$ as follows:
\begin{equation}
    \tilde{x}=\lambda \cdot x + (1-\lambda) \cdot x', \enspace \tilde{y} = \lambda \cdot y + (1-\lambda) \cdot y'.
\end{equation}
Verma \etal~\cite{verma2019manifold} extend \textit{mixup} to \textit{manifold mixup} that leverages semantic interpolations as additional training signal, and obtain neural networks with smoother decision boundaries at multiple levels
of representation. Yun \etal \cite{yun2019cutmix} further propose \textit{cutmix} where the patches are cut and pasted among training images and the ground truth labels are also mixed proportionally to the area of patches. 

As a powerful data augmentation strategy, \textit{mixup} has also been used to enhance the robustness of deep models. Lamb \etal~\cite{lamb2019interpolated} propose \textit{interpolated adversarial training} (IAT) that adopts the adversarial examples processed by \textit{mixup} or \textit{manifold mixup} for training. Pang \etal~\cite{pang2020mixup} propose \textit{mixup inference} (MI) by mixing the input with other random clean sample for inference. Lee \etal \cite{lee2020adversarial} propose \textit{adversarial vertex mixup} (AVM) by mixing the clean example and adversarial example to enhance the robustness of PGD adversarial training \cite{madry2018pgd}. Laugros \etal~\cite{laugros2020addressing} combine \textit{mixup} and \textit{targeted labeling adversarial training} (TLAT) that interpolates the target labels of adversarial examples with the ground-truth labels.

\section{Methodology}
In this section, we first provide details of several adversarial attacks for enhancing the transferability 
to which our method is most related. Then 
we introduce the proposed \textit{Admix} attack method and highlight the difference between the proposed \textit{admix} operation and the 
existing \textit{mixup}~\cite{zhang2018mixup} 
operation
designed for standard 
training. 

\subsection{Attacks for Enhancing the Transferability} 

Let $\mathcal{X}$ be the set of all digital images under consideration for a given learning task, $\mathcal{Y}\in \mathbb{R}$ be the output label space and $\mathcal{B}_\epsilon(x) = \{\bar{x} : \|x-\bar{x}\|_p \leq \epsilon\}$ denote the $\ell_p$-norm ball centered at $x$ with radius $\epsilon$. Given a classifier $f(x;\theta):x \in \mathcal{X} \rightarrow y \in \mathcal{Y}$ that outputs label $y$ for the prediction of input $x$ with model parameters $\theta$, the goal of adversarial attack is to seek an example $x^{adv} \in \mathcal{B}_\epsilon(x)$ that misleads the target classifier $f(x;\theta)\neq f(x^{adv}; \theta)$. To align with previous works, we focus on $\ell_\infty$-norm in this work.

\textbf{Fast Gradient Sign Method (FGSM)}~\cite{goodfellow2015FGSM} crafts adversarial example by adding perturbation in the gradient direction of the loss function $J(x,y;\theta)$ as follows:
\begin{equation*}
    x^{adv} = x + \epsilon \cdot \text{sign}(\nabla_x J(x,y;\theta)),
\end{equation*}
where $\text{sign}(\cdot)$ denotes the sign function and $\nabla_x J(x,y;\theta)$ is the gradient of the loss function \wrt $x$.

\textbf{Iterative Fast Gradient Sign Method (I-FGSM)}~\cite{kurakin2017IFGSM} is an iterative version of FGSM by adding a small perturbation with step size $\alpha$ in the gradient direction at each iteration:
\begin{equation*}
    x_{t+1}^{adv} = x_t^{adv} + \alpha \cdot \text{sign}(\nabla_{x_t^{adv}} J(x_t^{adv},y;\theta)), \enspace x_0^{adv} = x.
\end{equation*}

\textbf{Momentum Iterative Fast Gradient Sign Method (MI-FGSM)}~\cite{dong2018boosting} integrates the momentum term into I-FGSM and exhibits better transferability. The update procedure can be summarized as:
\begin{gather*}
    g_t = \mu \cdot g_{t-1} + \frac{\nabla_{x_t^{adv}}J(x_t^{adv},y,;\theta)}{\|\nabla_{x_t^{adv}}J(x_t^{adv},y,;\theta)\|_1}, \\
    x_{t+1}^{adv} = x_{t}^{adv} + \alpha \cdot \text{sign}(g_t).
\end{gather*}

\textbf{Diverse Input Method (DIM)}~\cite{xie2019improving} is the first input transformation based attack which firstly resizes the input image to an $r \times r \times 3$ image where $r$ is randomly sampled from $[299,330)$ with a given probability $p$ and pads the resized image into $330 \times 330 \times 3$. Then DIM feeds the transformed image to DNNs for gradient calculation.

\textbf{Translation-Invariant Method (TIM)}~\cite{dong2019evading} calculates the average gradient on a set of translated images for the update. To further improve the efficiency, TIM approximately calculates the gradient by convolving the gradient of the untranslated image with a predefined kernel matrix instead of computing the gradient on a set of images.

\textbf{Scale-Invariant Method (SIM)}~\cite{lin2020nesterov} discovers the scale invariance property of DNNs and calculates the average gradient over the scaled copies of the input for update:
\begin{equation*}
    \bar{g}_{t+1} = \frac{1}{m} \sum_{i=0}^{m-1} \nabla_{x_{t}^{adv}} (J(x_{t}^{adv} / 2^i, y; \theta)),
\end{equation*}
where $m$ is the number of copies.

\subsection{The Admix Attack Method} 
Lin \etal~\cite{lin2020nesterov} analogize the adversary generation process to the neural model training process and the transferability of crafted adversarial example could be equivalent to the generalization of the trained model. Under such perspective, the input transformation could be treated as data augmentation. Various input transformations have been proposed that could boost the adversarial transferability, however, we observe that all the existing transformations are applied on the single input image. On the other hand, we observe that for standard training, \textit{mixup}, which is a powerful data augmentation strategy by interpolating two randomly sampled examples, can effectively improve the model generalization~\cite{zhang2018mixup, tokozume2018between, yun2019cutmix}.
This raises an intriguing question, \textit{could we improve the attack transferability by adopting information from other images for the gradient calculation?}

However, as shown in Table \ref{tab:mixup}, we find that directly applying \textit{mixup} for the gradient calculation improves the transferability of crafted adversaries slightly but degrades the attack performance significantly under white-box setting. 
The main reason might be two-fold. First, there is no difference between $x$ and $x'$ for the \textit{mixup} which might adopt too much information from the add-in image $x'$ for the gradient calculation of the input $x$ and thus provide incorrect direction for update. Second, \textit{mixup} also mixes the labels which introduces the gradient of other category for update when $x$ and $x'$ are not in the same category.

\begin{algorithm}[tb]
    \algnewcommand\algorithmicinput{\textbf{Input:}}
    \algnewcommand\Input{\item[\algorithmicinput]}
    \algnewcommand\algorithmicoutput{\textbf{Output:}}
    \algnewcommand\Output{\item[\algorithmicoutput]}
    \caption{The \textit{Admix} Attack Algorithm}
    \label{alg:aam}
	\begin{algorithmic}[1]
		\Input A classifier $f$ with loss function $J$ and a benign example $x$ with ground-truth label $y$
		\Input The maximum perturbation $\epsilon$, number of iterations $T$ and decay factor $\mu$
		\Input The number of admixed copies $m_1$ and sampled images $m_2$, and the strength of sampled image $\eta$
        \Output An adversarial example $x^{adv} \in \mathcal{B}_{\epsilon}(x)$
		\State $\alpha = \epsilon/T$; $g_0 = 0$; $\bar{g}_0 = 0$; $x_0^{adv}=x$
		\For{$t = 0 \rightarrow T-1$}:
		    \State Randomly sample a set $X'$ of $m_2$ images from another category  
		    \State Calculate the average gradient $\bar{g}_{t+1}$ by Eq. \eqref{eq:aam}
		    \State Update the enhanced momentum $g_{t}$:
		    \begin{equation*}
		        g_{t+1} = \mu \cdot g_{t} + \frac{\bar{g}_{t+1}}{\|\bar{g}_{t+1}\|_1}
		    \end{equation*}
		    \State Update $x_{t+1}^{adv}$ by applying the gradient sign: 
		    \begin{equation*}
		        x^{adv}_{t+1} = x_t^{adv} + \alpha \cdot \text{sign}(g_{t+1})
		    \end{equation*}
		\EndFor
        \State \Return $x^{adv}=x_{T}^{adv}$.
	\end{algorithmic} 
\end{algorithm}

In order to utilize the information of images from other category without harming the white-box attack performance, we propose \textit{admix} operation that admixes two images in a master and slave manner. Specifically, we takes the original image $x$ as the primary image and admixes it with a secondary image $x'$ randomly picked from other category: 
\begin{equation}
    \tilde{x} = \gamma \cdot x + \eta' \cdot x' = \gamma \cdot (x + \eta \cdot x'),
    \label{eq:admix}
\end{equation}
where $\eta = \eta' / \gamma$, $\gamma \in [0,1]$ and $\eta' \in [0, \gamma)$ control the portion of the original image and the randomly sampled image in the admixed image respectively. In this way, we can assure that the secondary image $x'$ always occupies a smaller portion in $\tilde{x}$. Note that we do not mix the labels, but instead use the original label of $x$ for $\tilde{x}$.

\begin{figure}
    \centering
    \includegraphics[width=\linewidth]{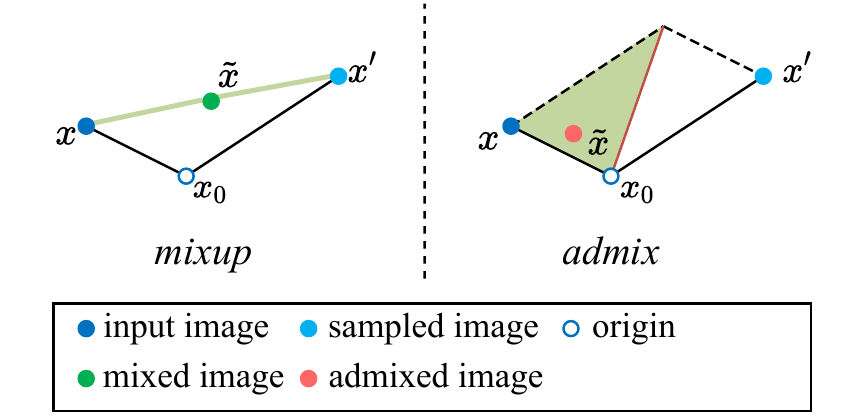}
    \caption{Illustration of the mechanisms in the input space of \textit{mixup} and \textit{admix}. $x$ denotes the input image and $x'$ the randomly sampled image. $x_0$ denotes the origin where all pixel values are 0s and $\tilde{x}$ is a possible transformed image. The green line and green triangle denotes all the possible transformed images by \textit{mixup} and \textit{admix}, respectively.}
    \label{fig:mixup_admix}
    \vspace{-1em}
\end{figure}
With the above analysis, we propose an \textit{Admix} attack method to improve the attack transferability, which calculates the average gradient on a set of admixed images \{$\tilde{x}$\} of the input $x$ by changing the value of $\gamma$ or picking the add-in image $x'$ from different categories in Eq. \eqref{eq:admix}. 
\begin{equation}
    \bar{g}_{t+1} = \frac{1}{m_1 \cdot m_2} \sum_{x'\in X'} \sum_{i=0}^{m_1-1} \nabla_{x_t^{adv}} J(\gamma_i \cdot (x_t^{adv}+\eta \cdot x'), y; \theta),
    \label{eq:aam}
\end{equation}
where $m_1$ is the number of admixed images for each $x'$ and $X'$ denotes the set of $m_2$ randomly sampled images from other categories. Note that when $\eta = 0$, 
\textit{Admix} will degenerate 
to SIM~\cite{lin2020nesterov}. The proposed \textit{Admix} could be integrated with any gradient-based attacks and other input transformation methods except for SIM. We summarize the algorithm of \textit{Admix} integrated into MI-FGSM (denoted as \textit{Admix} without ambiguity in the following) in Algorithm \ref{alg:aam}.

\begin{table*}
\begin{center}
\scalebox{0.9}{
\begin{tabular}{|l|c|ccccccc|}
\hline
Model & Attack & Inc-v3 & Inc-v4 & IncRes-v2 & Res-101 & Inc-v3$_{ens3}$ & Inc-v3$_{ens4}$ & IncRes-v2$_{ens}$\\
\hline\hline
\multirow{5}{*}{Inc-v3}& DIM & ~~99.0* & 64.3 & 60.9 & 53.2 & 19.9 & 18.3 &  ~~9.3\\
& TIM & \textbf{100.0}* & 48.8 & 43.6 & 39.5 & 24.8 & 21.3 & 13.2\\
& SIM & \textbf{100.0}* & 69.4 & 67.3 & 62.7 & 32.5 & 30.7 & 17.3\\
& \textit{Admix} & \textbf{100.0}* & \textbf{82.6} & \textbf{80.9} & \textbf{75.2} & \textbf{39.0} & \textbf{39.2} & \textbf{19.2} \\
\hline
\multirow{5}{*}{Inc-v4}& DIM & 72.9 & ~~97.4* & 65.1 & 56.5 & 20.2 & 21.1 & 11.6\\
& TIM & 58.6 & \textbf{~~99.6}* & 46.5 & 42.3 & 26.2 & 23.4 & 17.2\\
& SIM & 80.6 & \textbf{~~99.6}* & 74.2 & 68.8 & 47.8 & 44.8 & 29.1\\
& \textit{Admix} & \textbf{87.8} & ~~99.4* & \textbf{83.2} & \textbf{78.0} & \textbf{55.9} & \textbf{50.4} & \textbf{33.7}\\
\hline
\multirow{5}{*}{IncRes-v2}& DIM & 70.1 & 63.4 & ~~93.5* & 58.7 & 30.9 & 23.9 & 17.7\\
& TIM & 62.2 & 55.4 & ~~97.4* & 50.5 & 32.8 & 27.6 & 23.3\\
& SIM & 84.7 & 81.1 & ~~99.0* & 76.4 & 56.3 & 48.3 & 42.8\\
& \textit{Admix} & \textbf{89.9} & \textbf{87.5} & ~~\textbf{99.1}* & \textbf{81.9} & \textbf{64.2} & \textbf{56.7} & \textbf{50.0} \\
\hline
\multirow{5}{*}{Res-101}& DIM & 75.8 & 69.5 & 70.0 & ~~98.0* & 35.7 & 31.6 & 19.9\\
& TIM & 59.3 & 52.1 & 51.8 & ~~99.3* & 35.4 & 31.3 & 23.1\\
& SIM & 75.2 & 68.9 & 69.0 & ~~\textbf{99.7}* & 43.7 & 38.5 & 26.3\\
& \textit{Admix} & \textbf{85.4} & \textbf{80.8} & \textbf{79.6} & \textbf{~~99.7}* & \textbf{51.0} & \textbf{45.3} & \textbf{30.9}\\
\hline
\end{tabular}
}
\vspace{-0.3em}
\caption{Attack success rates (\%) on seven models under single model setting with various single input transformations. The adversaries are crafted on Inc-v3, Inc-v4, IncRes-v2 and Res-101 model respectively. * indicates white-box attacks.}
\label{tab:single_transformation}
\end{center}
\vspace{-1.5em}
\end{table*}
\subsection{Differences between Admix and Mixup}
\label{sec:diff}
For the two operations, \textit{admix} and \textit{mixup}~\cite{zhang2018mixup}, they both generate 
a mixed image from an image pair, $x$ and $x'$. Here we summarize their differences as follows:
\begin{itemize}[leftmargin=*]
    \setlength{\itemsep}{0pt}
    \setlength{\parsep}{0.1pt}
    \setlength{\parskip}{0pt}
    \item The goal of \textit{mixup} is to improve the generalization of the trained DNNs while \textit{admix} aims to generate more transferable adversarial examples.    
    \item The \textit{mixup} treats $x$ and $x'$ equally and also mixes the label of $x$ and $x'$. In contrast, \textit{admix} treats $x$ as the primary component and combines a small portion of $x'$, at the same time maintains the label of $x$.
    \item As depicted in Figure \ref{fig:mixup_admix}, \textit{mixup} linearly interpolates $x$ and $x'$ while \textit{admix} does not have such constraint, leading to more diversed transformed images.
\end{itemize}

\section{Experiments}
In this section, to validate the effectiveness of the proposed approach, we conduct extensive empirical evaluations on the standard ImageNet dataset~\cite{russakovsky2015imagenet}.


\subsection{Experimental Setup}
\label{sec:setup}
\textbf{Dataset.} We evaluate the proposed method on 1,000 images pertaining to 1,000 categories that are randomly sampled from the ILSVRC 2012 validation set~\cite{russakovsky2015imagenet} provided by Lin \etal~\cite{lin2020nesterov}. 

\textbf{Baselines.} We adopt three competitive input transformations as our baselines, \ie DIM~\cite{xie2019improving}, TIM~\cite{dong2019evading} and SIM~\cite{lin2020nesterov} and their combinations, denoted as SI-TIM, SI-DIM and SI-TI-DIM, respectively. All the input transformations are integrated into MI-FGSM~\cite{dong2018boosting}. 

\textbf{Models.} We study four popular normally trained models, \ie Inception-v3 (Inc-v3)~\cite{szegedy2016inceptionv3}, Inception-v4 (Inc-v4), Inception-Resnet-v3 (IncRes-v3)~\cite{szegedy2017inception} and Resnet-v2-101 (Res-101)~\cite{he2016resnet} and three ensemble adversarially trained models, \ie ens3-adv-Inception-v3 (Inc-v3$_{ens3}$), ens4-Inception-v3 (Inc-v3$_{ens4}$), ens-adv-Inception-ResNet-v2 (IncRes-v2$_{ens}$)~\cite{tramer2018ensemble}. In the following, we simply call the last three models as adversarially trained models without ambiguity. To further show the effectiveness of \textit{Admix}, we consider nine extra advanced defense models that are shown to be robust against black-box attacks on ImageNet dataset, namely HGD~\cite{liao2018defense}, R\&P~\cite{xie2018mitigating}, NIPS-r3\footnote{\url{https://github.com/anlthms/nips-2017/tree/master/mmd}}, Bit-Red~\cite{xu2018BitReduction}, FD~\cite{liu2019FD}, JPEG~\cite{guo2018countering}, RS~\cite{cohen2019certified}, ARS~\cite{salman2019provably} and NRP~\cite{naseer2020NRP}.

\textbf{Attack setting.} We follow the attack settings in~\cite{dong2018boosting} with the maximum perturbation of $\epsilon=16$, number of iteration $T=10$, step size $\alpha=1.6$ and the decay factor for MI-FGSM $\mu=1.0$. We adopt the Gaussian kernel with size $7 \times 7$ for TIM, the transformation probability $p=0.5$ for DIM, and the number of copies $m=5$ for SIM. For a fair comparison, we set $m_1 = 5$ with $\gamma_i = 1/2^i$ as in SIM, and randomly sample $m_2 = 3$ images with $\eta = 0.2$ for \textit{Admix}.

\subsection{Evaluation on Single Input Transformation}
\label{sec:single_transformation}
We first evaluate the attack performance of various single input transformations, namely DIM, TIM, SIM and the proposed \textit{Admix} attack. We craft adversaries on four normally trained networks respectively and test them on all the seven considered models. The attack success rates, \ie the misclassification rates of the corresponding models with adversaries as the inputs, are shown in Table~\ref{tab:single_transformation}. The models we attack are on rows and the models we test are on columns.

\begin{table*}
\begin{center}
\begin{subtable}{1.0\textwidth}
\centering
\scalebox{0.9}{
\begin{tabular}{|l|c|ccccccc|}
\hline
Model & Attack & Inc-v3 & Inc-v4 & IncRes-v2 & Res-101 & Inc-v3$_{ens3}$ & Inc-v3$_{ens4}$ & IncRes-v2$_{ens}$\\
\hline\hline
\multirow{2}{*}{Inc-v3}& SI-DIM & ~~98.9* & 85.0 & 81.3 & 76.3 & 48.0 & 45.1 & 24.9\\
& \textit{Admix}-DIM & ~~\textbf{99.8}* & \textbf{90.5} & \textbf{87.7} & \textbf{83.5} & \textbf{52.2} & \textbf{49.9} & \textbf{28.6}\\
\hline
\multirow{2}{*}{Inc-v4}& SI-DIM & 89.3 & ~~98.8* & 85.6 & 79.9 & 58.4 & 55.2 & 39.3\\
& \textit{Admix}-DIM  & \textbf{93.0} & ~~\textbf{99.2}* & \textbf{89.7} & \textbf{85.2} & \textbf{62.4} & \textbf{60.3} & \textbf{39.7}\\
\hline
\multirow{2}{*}{IncRes-v2}& SI-DIM & 87.9 & 85.1 & ~~97.5* & 82.9 & 66.0 & 59.3 & 52.2\\
& \textit{Admix}-DIM  & \textbf{90.2} & \textbf{88.4} & ~~\textbf{98.0}* & \textbf{85.8} & \textbf{70.5} & \textbf{63.7} & \textbf{55.3}\\
\hline
\multirow{2}{*}{Res-101}& SI-DIM & 87.9 & 83.4 & 84.0 & ~~98.6* & 63.5 & 57.5 & 42.0\\
& \textit{Admix}-DIM  & \textbf{91.9} & \textbf{89.0} & \textbf{89.6} & ~~\textbf{99.8}* & \textbf{69.7} & \textbf{62.3} & \textbf{46.6}\\
\hline
\end{tabular}
}
\vspace{-0.4em}
\caption{Attack success rates (\%) on seven models by SIM and \textit{Admix} integrated with \textbf{DIM}.}
\end{subtable}

\vspace{0.3em}

\begin{subtable}{1.0\textwidth}
\centering
\scalebox{0.9}{
\begin{tabular}{|l|c|ccccccc|}
\hline
Model & Attack & Inc-v3 & Inc-v4 & IncRes-v2 & Res-101 & Inc-v3$_{ens3}$ & Inc-v3$_{ens4}$ & IncRes-v2$_{ens}$\\
\hline\hline
\multirow{2}{*}{Inc-v3}& SI-TIM & \textbf{100.0}* & 71.8 & 68.6 & 62.2 & 48.2 & 47.4 & 31.3\\
& \textit{Admix}-TIM  & \textbf{100.0}* & \textbf{83.9} & \textbf{80.4} & \textbf{74.4} & \textbf{59.1} & \textbf{57.9} & \textbf{39.2}\\
\hline
\multirow{2}{*}{Inc-v4}& SI-TIM & 78.2 & ~~99.6* & 71.9 & 66.1 & 58.6 & 55.4 & 45.1\\
& \textit{Admix}-TIM  & \textbf{87.4} & ~~\textbf{99.7}* & \textbf{82.3} & \textbf{77.0} & \textbf{68.1} & \textbf{65.3} & \textbf{53.1}\\
\hline
\multirow{2}{*}{IncRes-v2}& SI-TIM & 84.5 & 82.2 & ~~\textbf{98.8}* & 77.4 & 71.6 & 64.7 & 61.0\\
& \textit{Admix}-TIM  & \textbf{90.2} & \textbf{88.2} & ~~98.6* & \textbf{83.9} & \textbf{78.4} & \textbf{73.6} & \textbf{70.0}\\
\hline
\multirow{2}{*}{Res-101}& SI-TIM & 74.2 & 69.9 & 70.2 & ~~\textbf{99.8}* & 59.5 & 54.5 & 42.8\\
& \textit{Admix}-TIM  & \textbf{83.2} & \textbf{78.9} & \textbf{80.7} & ~~99.7* & \textbf{67.0} & \textbf{62.5} & \textbf{52.8}\\
\hline
\end{tabular}
}
\vspace{-0.4em}
\caption{Attack success rates (\%) on seven models by SIM and \textit{Admix} integrated with \textbf{TIM}.}
\end{subtable}

\vspace{0.3em}

\begin{subtable}{1.0\textwidth}
\centering
\scalebox{0.9}{
\begin{tabular}{|l|c|ccccccc|}
\hline
Model & Attack & Inc-v3 & Inc-v4 & IncRes-v2 & Res-101 & Inc-v3$_{ens3}$ & Inc-v3$_{ens4}$ & IncRes-v2$_{ens}$\\
\hline\hline

\multirow{2}{*}{Inc-v3}& SI-TI-DIM & ~~99.1* & 83.6 & 80.8 & 76.7 & 65.2 & 63.3 & 46.5\\
& \textit{Admix}-TI-DIM & ~~\textbf{99.9}* & \textbf{89.0} & \textbf{87.0} & \textbf{83.1} & \textbf{72.2} & \textbf{71.1} & \textbf{52.4}\\
\hline

\multirow{2}{*}{Inc-v4}& SI-TI-DIM & 87.9 & ~~98.7* & 83.0 & 77.7 & 72.4 & 68.2 & 57.5\\
& \textit{Admix}-TI-DIM  & \textbf{90.4} & ~~\textbf{99.0}* & \textbf{87.3} & \textbf{82.0} & \textbf{75.3} & \textbf{71.9} & \textbf{61.6}\\
\hline

\multirow{2}{*}{IncRes-v2}& SI-TI-DIM & 88.8 & 86.8 & ~~\textbf{97.8}* & 83.9 & 78.7 & 74.2 & 72.3\\
& \textit{Admix}-TI-DIM  & \textbf{90.1} & \textbf{89.6} & ~~97.7* & \textbf{85.9} & \textbf{82.0} & \textbf{78.0} & \textbf{76.3}\\
\hline

\multirow{2}{*}{Res-101}& SI-TI-DIM & 84.7 & 82.2 & 84.8 & ~~99.0* & 75.8 & 73.5 & 63.4\\
& \textit{Admix}-TI-DIM  & \textbf{91.0} & \textbf{87.7} & \textbf{89.2} & ~~\textbf{99.9}* & \textbf{81.1} & \textbf{77.4} & \textbf{70.1} \\
\hline
\end{tabular}
}
\vspace{-0.3em}
\caption{Attack success rates (\%) on seven models by SIM and \textit{Admix} integrated with \textbf{TI-DIM}.}
\end{subtable}
\vspace{-0.4em}
\caption{Attack success rates (\%) on seven models under single model setting with various combined input transformations. The adversaries are crafted on Inc-v3, Inc-v4, IncRes-v2 and Res-101 model respectively. * indicates white-box attacks.}
\label{tab:combination}
\end{center} 
\vspace{-1em}
\end{table*}

We can see that TIM exhibits the weakest transferability on normally trained models among four input transformations, but outperforms DIM on adversarially trained models. SIM achieves better transferability than DIM and TIM on both normally trained models and adversarially trained models. Compared with the three competitive baselines, \textit{Admix} achieves much better transferability on all the models and maintains high attack success rates under white-box setting. For instance, 
both \textit{Admix} and SIM achieve the attack success rates of 100\% for white-box attack on Inc-v3 model, however for black-box attack, \textit{Admix} achieves the attack success rates of 82.6\% on Inc-v4 model and 39.0\% on Inc-v3$_{ens3}$ model 
while the powerful baseline SIM only achieves 
69.4\% on Inc-v4 and 30.7\% on Inc-v3$_{ens3}$.

\subsection{Evaluation on Combined Input Transformation}
\label{sec:combine_transformation}
Lin \etal~\cite{lin2020nesterov} show that combining SIM with TIM and DIM could further boost the transferability of the crafted adversaries. Here we evaluate the generalization of \textit{Admix} to other input transformations. Since SIM is a special case of \textit{Admix}, we compare the attack success rates of TIM and DIM integrated with SIM and \textit{Admix}, denoted as SI-DIM, SI-TIM, SI-TI-DIM, \textit{Admix}-DIM, \textit{Admix}-TIM and \textit{Admix}-TI-DIM respectively. We summarize the results in Table \ref{tab:combination}.

In general, the transformations combined with \textit{Admix} achieves better transferability than the ones combined with SIM on all models. Taking the adversaries crafted on Inc-v3 model for example, \textit{Admix}-DIM outperforms SI-DIM with a clear margin of 4\% $\sim$ 7\%, \textit{Admix}-TIM outperforms SI-TIM with a large margin of 8\% $\sim$ 12 \% and \textit{Admix}-TI-DIM outperforms SI-TI-DIM with a clear margin of 5\% $\sim$ 7\%. Such remarkable improvements demonstrate the high effectiveness of the proposed method by adopting extra information from other categories for the gradient calculation.

\subsection{Evaluation on Ensemble-model Attack}
\label{sec:ensemble_attack}
Liu \etal~\cite{liu2017delving}  have shown that attacking multiple models simultaneously can improve the transferability of the generated adversarial examples. To further demonstrate the efficacy of the proposed \textit{Admix}, we adopt the ensemble-model attack as in~\cite{dong2018boosting} by fusing the logit outputs of various models. The adversaries are generated on four normally trained models, namely Inc-v3, Inc-v4, IncRes-v2 and Res-101 using different input transformations and the integrated input transformations respectively. All the ensemble models are assigned with equal weights and we test the transferability of the adversaries on three adversarially trained models.

As shown in Table \ref{tab:ensemble_attack}, \textit{Admix} always achieves the highest attack success rates under both white-box and black-box settings no matter for single input transformation or integrated input transformation. Compared with single input transformation, \textit{Admix} achieves higher attack success rate that is at least 6.7\% higher than SIM, which achieves the best attack performance among the three baselines. When combined with DIM or TIM, \textit{Admix} outperforms the corresponding baseline with a clear margin of at least 4\%. When integrating \textit{Admix} into the combination of DIM and TIM, even though SI-TI-DIM exhibits great attack performance, \textit{Admix} can further improve the baseline for more than 2\% on three adversarially trained models. This convincingly demonstrates the high efficacy of adopting the information from other categories to enhance the transferability.

\begin{table*}
\begin{center}
\scalebox{0.9}{
\begin{tabular}{|c|ccccccc|}
\hline
Attack & Inc-v3 & Inc-v4 & IncRes-v2 & Res-101 & Inc-v3$_{ens3}$ & Inc-v3$_{ens4}$ & IncRes-v2$_{ens}$\\
\hline\hline
DIM & ~~99.4* & 97.4* & 94.9* & ~~99.8* & 58.1 & 51.1 & 34.9\\
TIM & ~~99.8* & 97.9* & 95.2* & ~~99.8* & 62.2 & 56.8 & 48.0\\
SIM & ~~99.9* & 99.3* & 98.3* & \textbf{100.0}* & 78.8 & 73.9 & 59.5\\
\textit{Admix} & \textbf{100.0}* & \textbf{99.6}* & \textbf{99.0}* & \textbf{100.0}* & \textbf{85.5} & \textbf{80.9} & \textbf{67.8}\\
\hline
SI-DIM & ~~\textbf{99.7}* & 98.9* & 97.7* & ~~99.9* & 85.2 & 83.3 & 71.3\\
\textit{Admix}-DIM & ~~\textbf{99.7}* & \textbf{99.5}* & \textbf{98.9}* & \textbf{100.0}* & \textbf{89.3} & \textbf{87.8} & \textbf{79.0}\\
\hline
SI-TIM & ~~\textbf{99.7}* & 99.0* & 97.6* & \textbf{100.0}* & 87.9 & 85.2 & 80.4\\
\textit{Admix}-TIM & ~~\textbf{99.7}* & \textbf{99.1}* & \textbf{98.1}* & \textbf{100.0}* & \textbf{91.8} & \textbf{89.7} & \textbf{85.8}\\
\hline
SI-TI-DIM & ~~99.6* & \textbf{98.9}* & 97.8* & ~~99.7* & 91.1 & 90.3 & 86.8\\
\textit{Admix}-TI-DIM & ~~\textbf{99.7}* & \textbf{98.9}* & \textbf{98.3}* & \textbf{100.0}* & \textbf{93.9} & \textbf{92.3} & \textbf{90.0} \\
\hline
\end{tabular}
}
\vspace{-0.5em}
\caption{Attack success rates (\%) on seven models under ensemble-model setting with various input transformations. The adversaries are crafted on the ensemble model, \ie Inc-v3, Inc-v4, IncRes-v2 and Res-101. * indicates white-box attacks.}
\label{tab:ensemble_attack}
\end{center}
\vspace{-1.3em}
\end{table*}
\begin{table*}
\begin{center}
\scalebox{0.9}{
\begin{tabular}{|c|cccccccccc|}
\hline
Attack & HGD & R\&P & NIPS-r3 & Bit-Red & FD & JPEG & RS & ARS & NRP & Average\\
\hline\hline
SI-TI-DIM & 91.4 & 88.0 & 90.0 & 75.7 & 88.0 & 93.2 & 69.2 & 46.4 & 77.1 & 79.9\\
\textit{Admix}-TI-DIM & \textbf{93.7} & \textbf{90.3} & \textbf{92.4} & \textbf{80.1} & \textbf{91.9} & \textbf{95.4} & \textbf{74.9} & \textbf{51.4} & \textbf{80.7} & \textbf{83.3}\\
\hline
\end{tabular}
}
\vspace{-0.5em}
\caption{Attack success rates (\%) on nine extra models with advanced defense by SI-TI-DIM and \textit{Admix}-TI-DIM respectively. The adversaries are crafted on the ensemble model, \ie Inc-v3, Inc-v4, IncRes-v2 and Res-101.}
\label{tab:advanced_defense}
\end{center}
\vspace{-1.em}
\end{table*}
\subsection{Evaluation on Advanced Defense Models}
\label{sec:advanced_attack}
To further show the effectiveness of our method, we consider nine extra advanced defense methods, \ie the top-3 defense methods in the NIPS 2017 competition (HGD (rank-1)~\cite{liao2018defense}, R\&P (rank-2)~\cite{xie2018mitigating} and NIPS-r3 (rank-3)), three popular input transformation based defenses (Bit-Red~\cite{xu2018BitReduction}, FD~\cite{liu2019FD} and JPEG~\cite{guo2018countering}), two certified defenses (RS~\cite{cohen2019certified} and ARS~\cite{salman2019provably}) and a powerful denoiser (NRP~\cite{naseer2020NRP}). The target model for Bit-Red, FG, JPEG and NRP is Inc-v3$_{ens3}$ and the other methods adopt the official models provided in the corresponding papers. From the above evaluations, SI-TI-DIM exhibits the best attack performance among all the baselines under the ensemble-model setting. Thus, we compare the proposed \textit{Admix}-TI-DIM 
with SI-TI-DIM under the ensemble-model setting as in Sec.~\ref{sec:ensemble_attack}. 

We can observe from Table \ref{tab:advanced_defense} that the proposed \textit{Admix}-TI-DIM achieves higher attack success rates on all the defense models than SI-TI-DIM and outperforms the baseline with a clear margin of 3.4\% on average. In general, \textit{Admix}-TI-DIM results in a larger margin compared with SI-TI-DIM when attacking more powerful defense methods. For instance, \textit{Admix}-TI-DIM outperforms the baseline more than 5.7\% and 5\% on the models trained by randomized smoothing (RS) and adversarially randomized smoothing (ARS) that both provide certified defense. 

\begin{figure*}[t]
\centering 
    \begin{minipage}[b]{.48\textwidth} 
        \begin{subfigure}{.48\textwidth}
          \centering 
          \includegraphics[width=\linewidth, height=5cm]{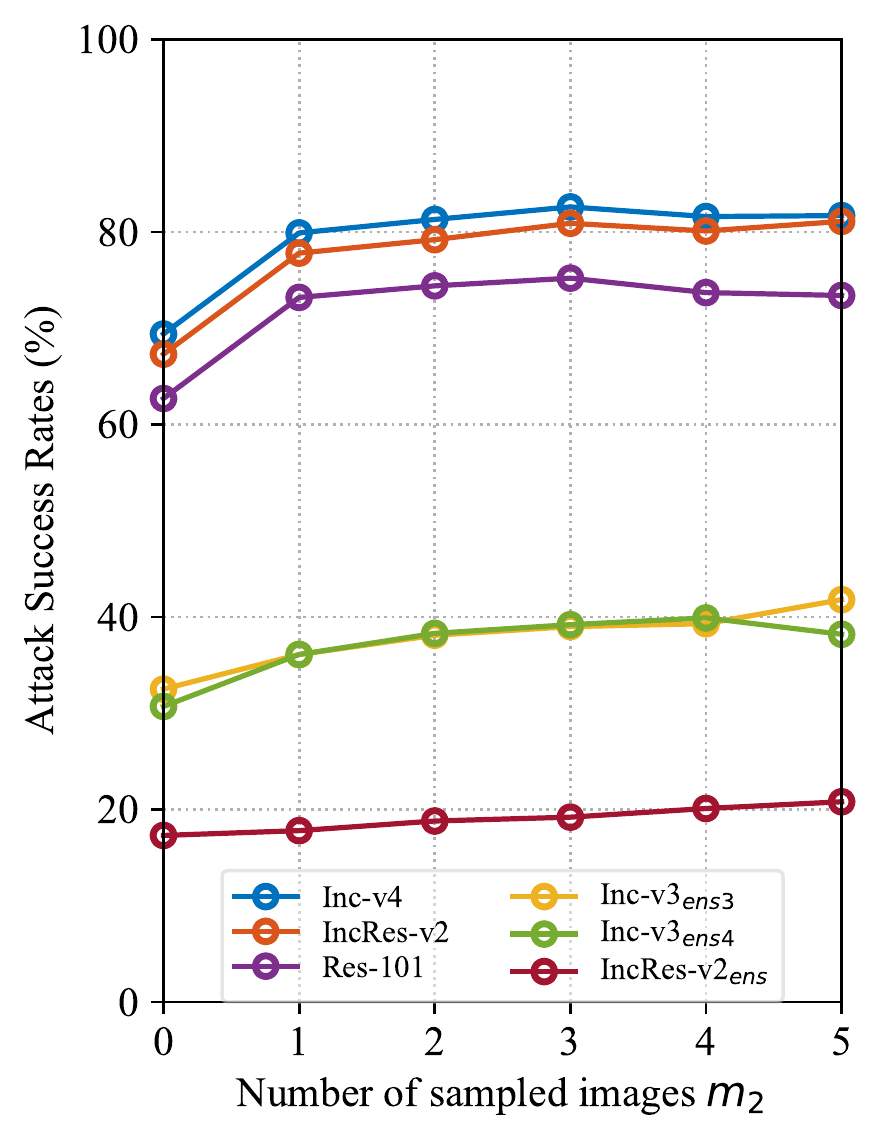}
          \vspace{-1.8em}
          \caption{\textit{Admix}}
          \label{fig:sampleInterval:MLNI-FGSM}
        \end{subfigure}
        \hspace{.1em}
        \begin{subfigure}{.48\textwidth} 
          \centering 
          \includegraphics[width=\linewidth, height=5cm]{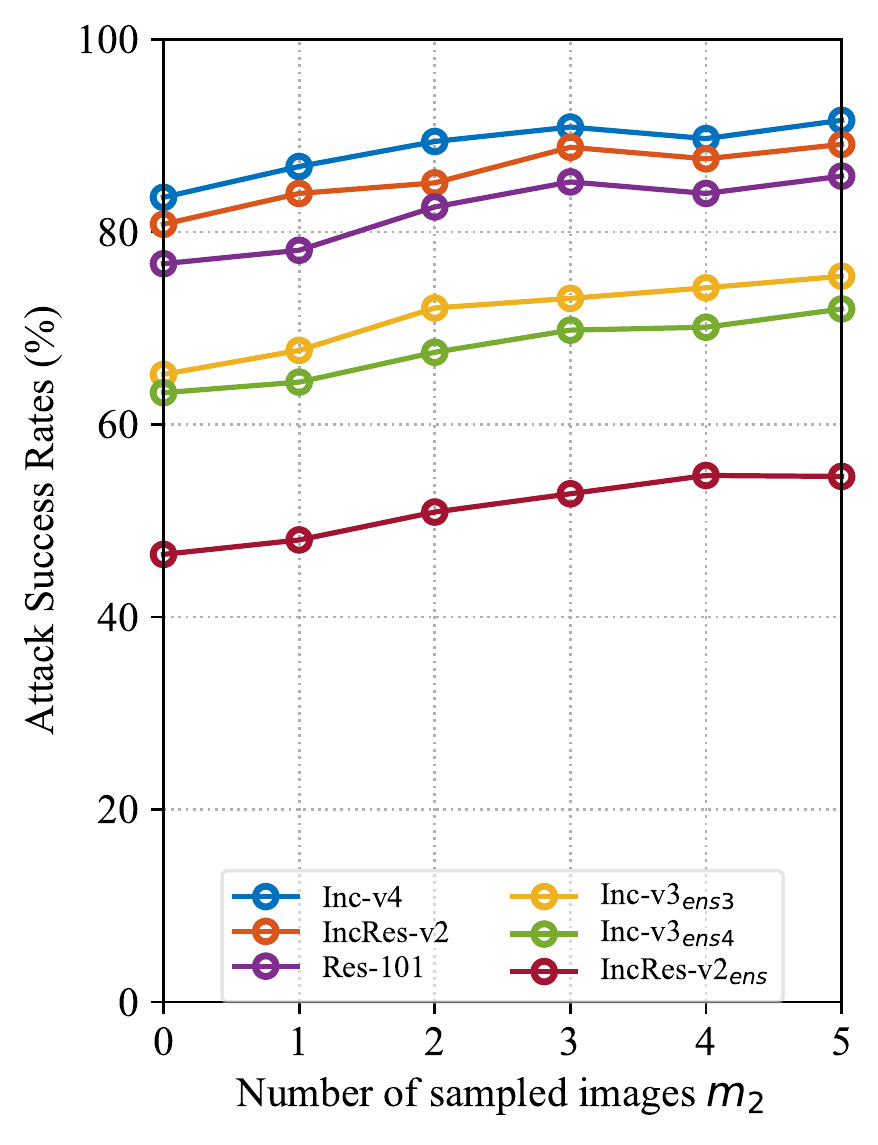}
          \vspace{-1.8em}
          \caption{\textit{Admix}-TI-DIM}
          \label{fig:sampleInterval:MLNI-DTS}
        \end{subfigure}%
    \vspace{-0.3em}
    \caption{Attack success rates (\%) on the other six models with adversaries crafted by \textit{Admix} and \textit{Admix}-TI-DIM on Inc-v3 model for various number of sampled images, $m_2$.}
    \label{fig:number}
    \end{minipage}
    \hspace{.3cm}
    \begin{minipage}[b]{0.48\textwidth} 
        \begin{subfigure}{.48\textwidth}
          \centering 
          \includegraphics[width=\linewidth, height=5cm]{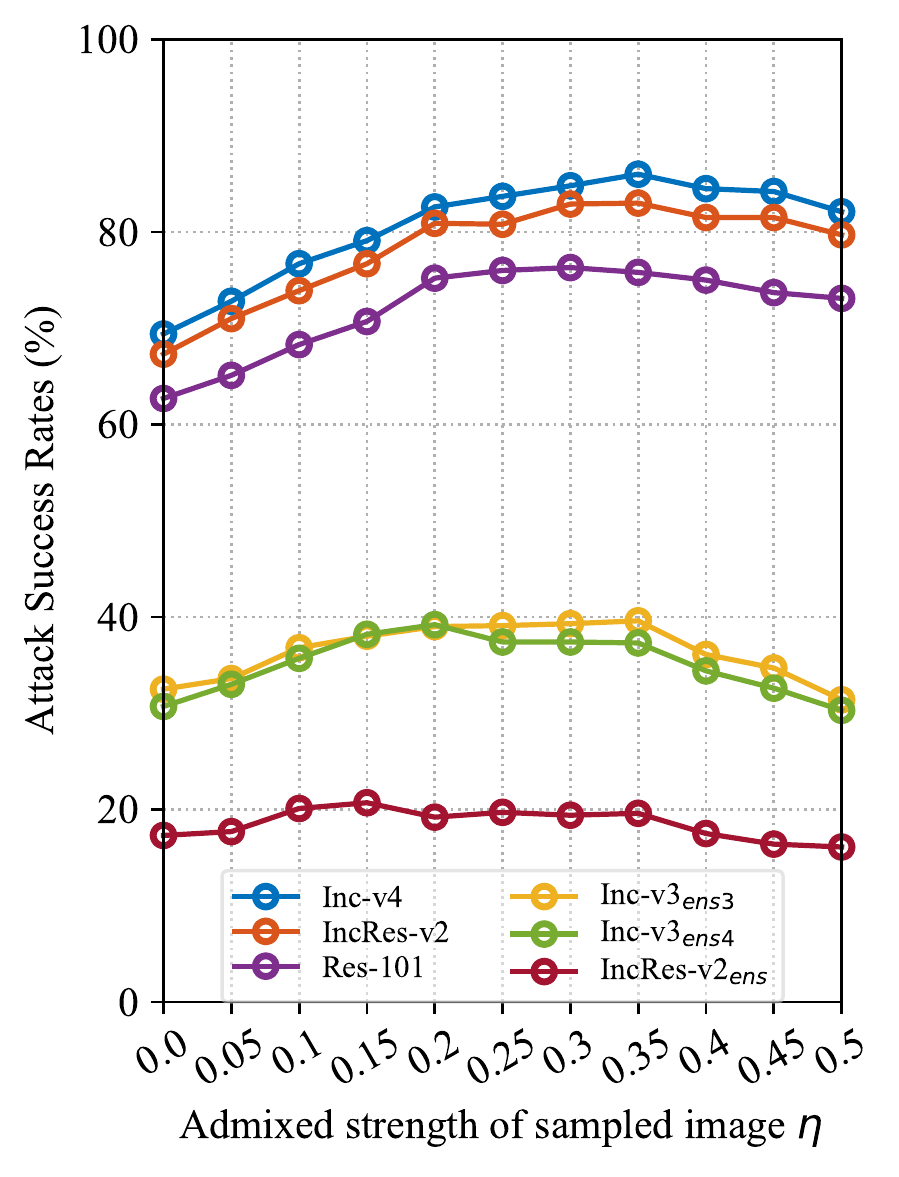}
          \vspace{-1.8em}
          \caption{\textit{Admix}}
          \label{fig:sampleNumber:MLNI-FGSM}
        \end{subfigure}
        \hspace{.1em}
        \begin{subfigure}{.48\textwidth} 
          \centering 
          \includegraphics[width=\linewidth, height=5cm]{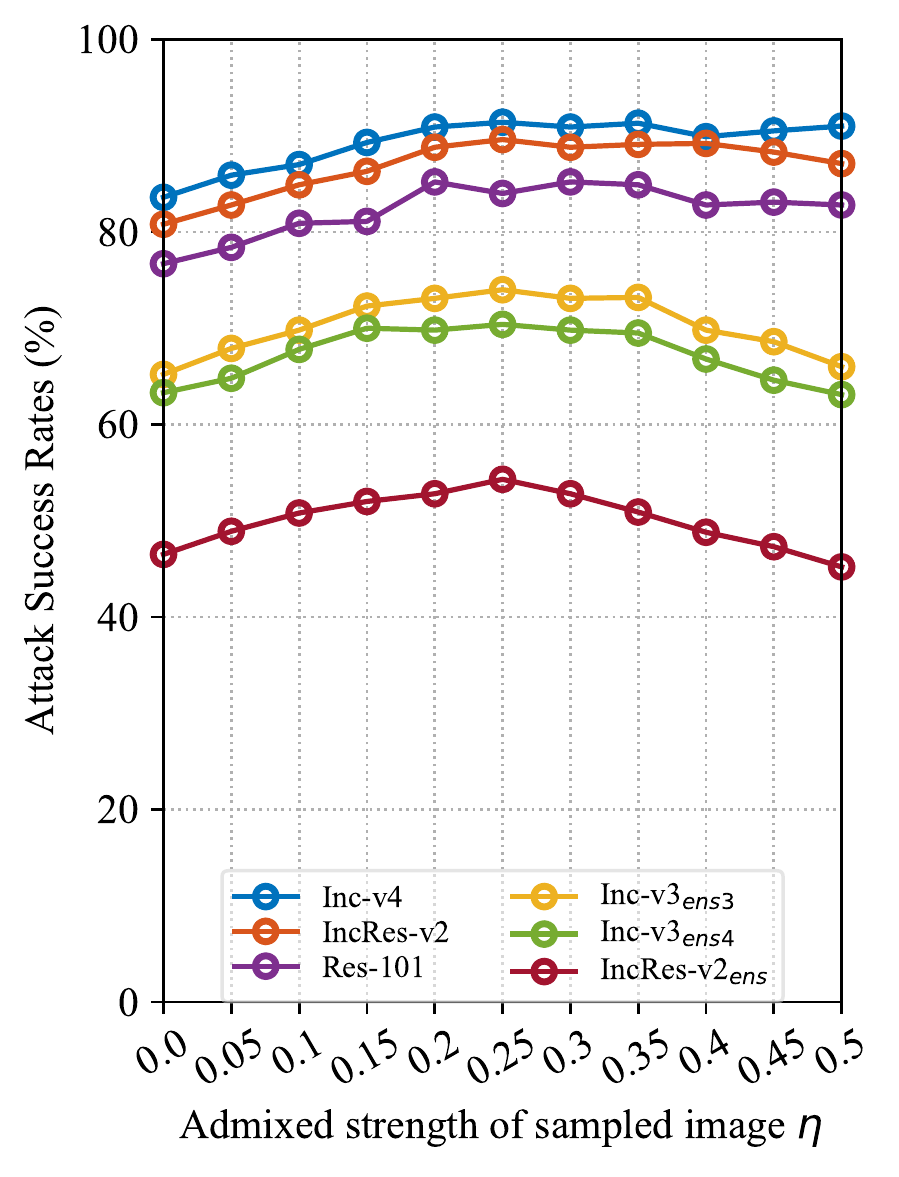}
          \vspace{-1.8em}
          \caption{\textit{Admix}-TI-DIM}
          \label{fig:eta}
        \end{subfigure}%
    \vspace{-0.3em}
    \caption{Attack success rates (\%) on the other six models with adversaries crafted by \textit{Admix} and \textit{Admix}-TI-DIM on Inc-v3 model for various strength of the sampled image, $\eta$.}
    \label{fig:eta}
    \end{minipage}
    \vspace{-1em}
\end{figure*}                                                                                                                                       

\subsection{Ablation Studies}
\label{sec:ablation}
For hyper-parameter $m_1$, we follow the setting of SIM~\cite{lin2020nesterov} for a fair comparison, and choose $m_1 = 5$. Here we conduct a series of ablation experiments to study the impact on \textit{Admix} and \textit{Admix}-TI-DIM on two other hyper-parameters, $m_2$ and $\eta$ used in experiments. 

\textbf{On the number of sampled images $x'$}. In Figure \ref{fig:number}, we report the attack success rates of \textit{Admix} and \textit{Admix}-TI-DIM for various values of $m_2$ with adversaries crafted on Inc-v3 model, where $\eta$ is fixed to 0.2. The attack success rates of \textit{Admix} are 100\% for all values of $m_2$ and that of \textit{Admix}-TI-DIM are at least 99.7\% under white-box setting. When $m_2=0$, \textit{Admix} and \textit{Admix}-TI-DIM degenerate to SIM and SI-TI-DIM respectively, and exhibit the weakest transferability. When $m_2 \leq 3$, the transferability on all models increases when we increase the value of $m_2$. When $m_2 > 3$, the transferability tends to decrease on normally trained models but still increases on adversarially trained models. Since a bigger value of $m_2$ indicates a higher computation cost, we set $m_2=3$ to balance the computational cost and attack performance.

\textbf{On the admixed strength of sampled image $x'$}. In Figure \ref{fig:eta}, we report the attack success rates of \textit{Admix} and \textit{Admix}-TI-DIM for various values of $\eta$ with adversaries crafted on Inc-v3 model, where $m_2$ is fixed to 3. The attack success rates of \textit{Admix} and \textit{Admix}-TI-DIM are at least 99.9\% and 99.7\% for various values of $\eta$ respectively under white-box setting. When $\eta=0$, \textit{Admix} and \textit{Admix}-TI-DIM also degenerate to SIM and SI-TI-DIM respectively, which exhibit the weakest transferability. When we increase $\eta$, the transferability increases rapidly and achieves the peak when $\eta=0.2$ for \textit{Admix} on adversarially trained models and $\eta=0.2$ or $\eta=0.25$  for \textit{Admix}-TI-DIM on all models. In general, we set $\eta=0.2$ for better performance. 

\begin{figure}
    \centering 
    \includegraphics[width=0.86\linewidth]{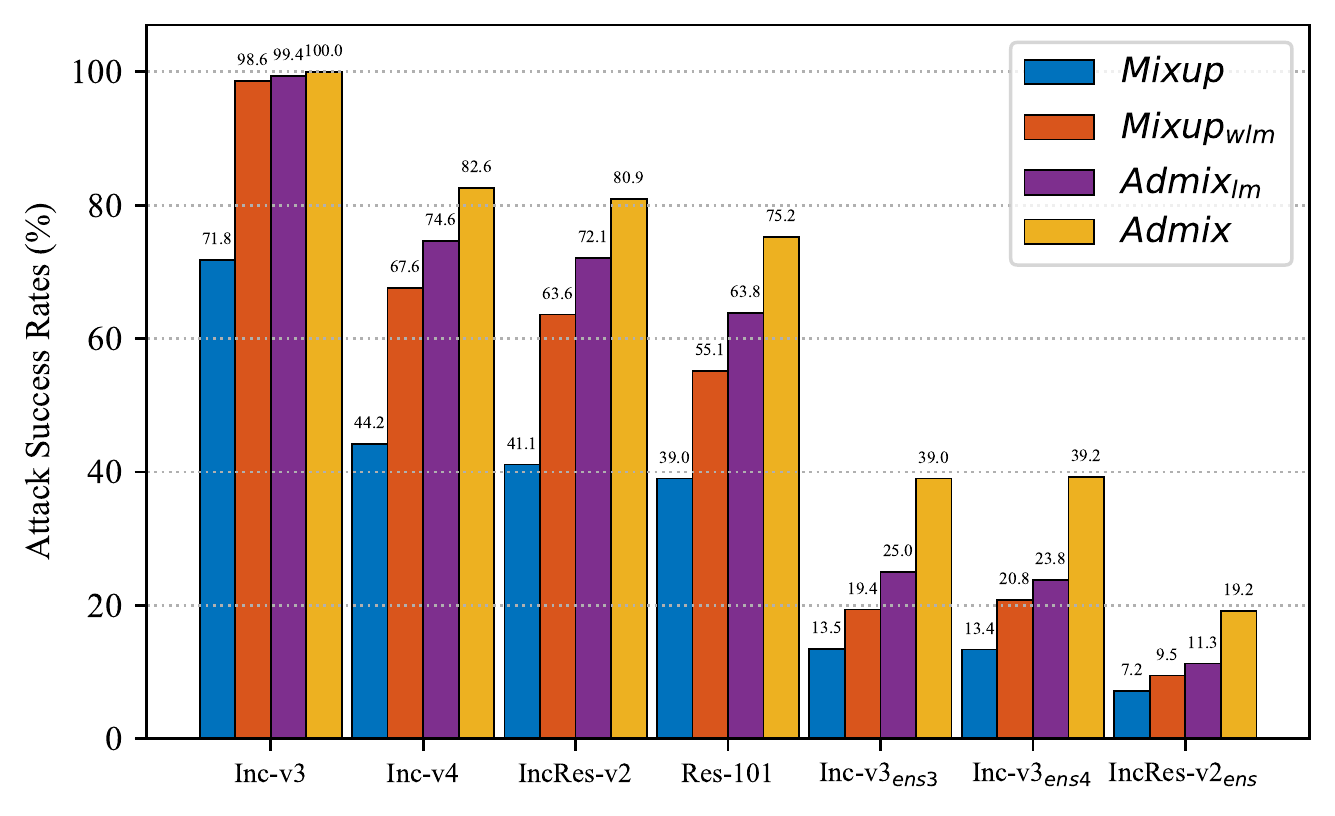}
    \vspace{-0.6em}
    \caption{Attack success rates (\%) on seven models with adversaries crafted by \textit{Mixup}, \textit{Mixup$_{wlm}$}, \textit{Admix$_{lm}$} and the proposed \textit{Admix} on Inc-v3 model.}
    \label{fig:mixup}
    \vspace{-1.2em}
\end{figure}
\subsection{Discussion}
\label{sec:discussion}
The \textit{admix} operation adds a small portion of the sampled image from other category to the original input but does not mix the labels. 
To verify the effectiveness of the two strategies, we evaluate the performance of \textit{mixup} without label mixing and \textit{admix} with label mixing for attacks, termed \textit{Mixup$_{wlm}$} and \textit{Admix$_{lm}$} respectively. 
As shown in Figure~\ref{fig:mixup}, \textit{Mixup$_{wlm}$} achieves better attack performance than \textit{Mixup}, which validates our hypothesis that the fact \textit{Mixup} mixes the labels and hence introduces the gradient of other category will 
weaken the attack performance.
We also see that \textit{Admix$_{lm}$} achieves higher attack success rates than \textit{Mixup} and \textit{Mixup$_{wlm}$}, highlighting the importance that the input image should be dominant in the mixed image. 
Thus, \textit{Admix} lets the input image be the dominant but does not mix the labels, and achieves the best attack performance.

\begin{figure}
    \centering 
    \includegraphics[width=0.86\linewidth]{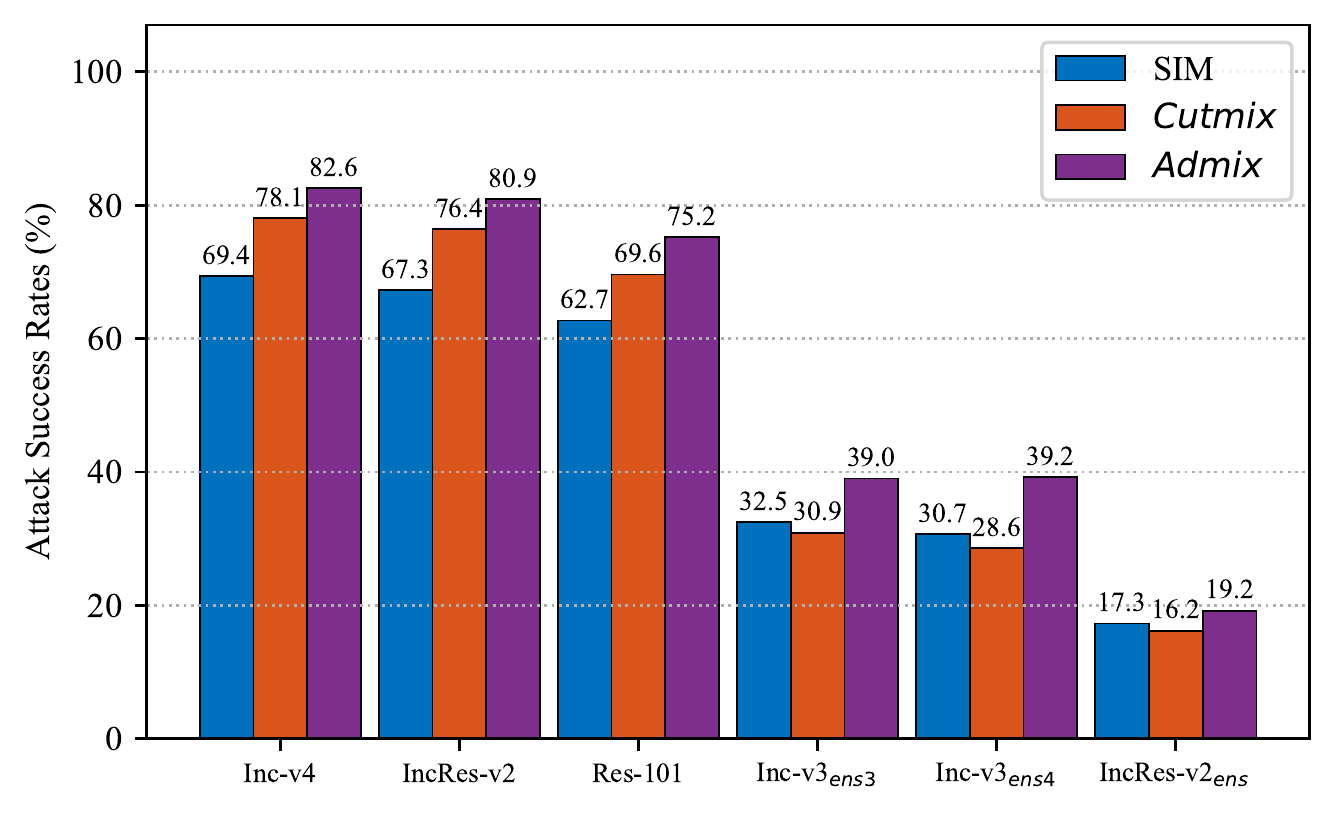}
    \vspace{-0.6em}
    \caption{Attack success rates (\%) on the other six models with adversaries crafted by SIM, \textit{Cutmix} and the proposed \textit{Admix} on Inc-v3 model.}
    \label{fig:cutmix}
    \vspace{-1.2em}
\end{figure}

We further provide a brief discussion on why \textit{Admix} helps craft more transferable adversarial examples. \textit{Admix} adds a small portion of the image from other class that moves the data point towards other class, \ie closer to the decision boundary. 
We hypothesize that \textit{Admix} utilizes the data points closer to the decision boundary for gradient calculation so that it could obtain more accurate update direction.
Similar strategy has also been used by NI-FGSM~\cite{lin2020nesterov} which looks ahead for the gradient calculation. To verify this hypothesis, we adopt \textit{Cutmix} input transformation~\cite{yun2019cutmix}, which randomly cuts a patch of input image and pastes a patch from another image designed for standard training, with the same procedure as \textit{Admix}.
Note that \textit{Cutmix} is not an interpolation of two images and does not guarantee the data point to be closer to the decision boundary. As shown in Figure~\ref{fig:cutmix}, \textit{Cutmix} exhibits better transferability on normally trained models but lower transferability on adversarially trained models when compared with SIM, and both \textit{Cutmix} and SIM exhibit much lower transferability than \textit{Admix}. This indicates that adopting information from other categories cannot always enhance the transferability and validate the hypothesis.


\section{Conclusion}
In this work, we propose a novel input transformation method called \textit{Admix} to boost the transferability of the crafted adversaries. Specifically, for each input images, we randomly sample a set of image from other categories and admix a minor portion for each sampled image into the original image to craft a set of diverse images but using the original label for the gradient calculation. 
Extensive evaluations demonstrate that the proposed \textit{Admix} attack method could achieve much better adversarial transferability than the existing competitive input transformation based attacks
while maintaining high attack success rates under white-box setting. 
In our opinion, the \textit{admix} operation is a new paradigm of data argumentation for adversarial learning in which the admixed images are closer to the decision boundary, offering more transferable adversaries. We hope our \textit{Admix} attack that adopts information from other categories will shed light on potential directions for adversarial attacks. 



\textbf{Acknowledgements:} This work is supported by National Natural Science Foundation (62076105) and Microsft Research Asia Collaborative Research Fund (99245180).

{\small
\bibliographystyle{ieee_fullname}
\bibliography{egbib}
}

\end{document}